%% file: main.tex
\documentclass{article}

\usepackage[dblblindworkshop,final]{neurips_2026}
\workshoptitle{SocialAgent: Second Workshop on Large Language Models for Social Reasoning and Simulation}

\usepackage[utf8]{inputenc}
\usepackage[T1]{fontenc}
\usepackage{amsmath}
\usepackage{amssymb}
\usepackage{booktabs}
\usepackage{float}
\usepackage{longtable}
\usepackage{graphicx}
\usepackage{microtype}
\usepackage{xcolor}
\usepackage{hyperref}
\usepackage{url}
\hypersetup{
  hidelinks,
  pdftitle={Distributionally-Verifiable Rewards for Human-Behavior Simulation},
  pdfauthor={Ruoyun Ma, Shuang Liu, Peiye Liu, Ao Qu, Baichuan Mo}
}
\usepackage{algorithm}
\usepackage{algpseudocode}

\title{Distributionally-Verifiable Rewards for
Human-Behavior Simulation}

\author{%
  Ruoyun Ma$^{1}$ \quad
  Shuang Liu$^{2}$ \quad
  Peiye Liu$^{3}$ \quad
  Ao Qu$^{4}$ \quad
  Baichuan Mo$^{1}$\thanks{Corresponding author: \href{mailto:bmo@tsinghua.edu.cn}{bmo@tsinghua.edu.cn}.} \\
  $^{1}$Tsinghua University \quad $^{2}$Carnegie Mellon University \\
  $^{3}$ByteDance \quad $^{4}$Massachusetts Institute of Technology
}

\begin{document}

\maketitle

\begin{abstract}
Large language models are increasingly used to simulate survey respondents in social-science research. A useful simulator should predict individual responses, reproduce population response distributions, and capture demographic differences. Common training approaches struggle to balance these goals: exact-match rewards encourage the model to concentrate on the most frequent human response, which is contradictory to the goal of distributional alignment. Meanwhile, aggregate distribution matching can look good even when demographic information is ignored.
To this end, we introduce Distributionally-Verifiable Rewards (DVR), a reinforcement-learning reward that uses demographic group distribution gaps to guide individual responses, designed by aligning policy generated frequencies with human response frequencies within each question-group margins.
We evaluate the method using the OpinionQA dataset and find that comparing with accuracy-only RLVR, DVR preserves per-response accuracy while reducing JSD by \(34.5\%\) in-distribution and \(14.6\%\) on held-out questions, with higher DPC point estimates on both splits.
Comparing KL-SFT with the demographics-blind marginal further shows that low JSD is insufficient to establish fidelity to population heterogeneity, highlighting that accuracy, distributional fit, and demographic structure should be evaluated together.

\end{abstract}

\section{Introduction}
% 0827 version
Large language models (LLMs) are increasingly used to simulate survey
respondents in social-science research
\citep{santurkar2023opinionqa,ahnert2026survey}. Unlike tasks evaluated against
a single correct label, faithful simulation must capture individual answers,
population response frequencies, and systematic demographic variation. It is
therefore a form of \emph{distributionally pluralistic alignment}: reflecting a
target population's distribution of perspectives rather than collapsing them
into one consensus \citep{sorensen2024roadmap}.

Current work on human-response simulation largely pursues two complementary objectives: improving individual-response prediction \citep{kolluri2025socsci,binz2025foundation} and aligning model outputs with population response distributions \citep{meister2025benchmarking,suh2025subpop,yao2025gdpo}. Reinforcement learning has shown promise for the former objective \citep{berdoz2026reasoning}; however, population-level discrepancies do not naturally decompose into rewards for individual completions, making existing distributional objectives difficult to integrate with general-purpose RL methods such as GRPO \citep{shao2024deepseekmath}. Since both individual accuracy and population-level fidelity matter, a useful method should expose their trade-off and ideally optimize both jointly.

\begin{figure}[t]
  \centering
  \includegraphics[width=0.97\linewidth]{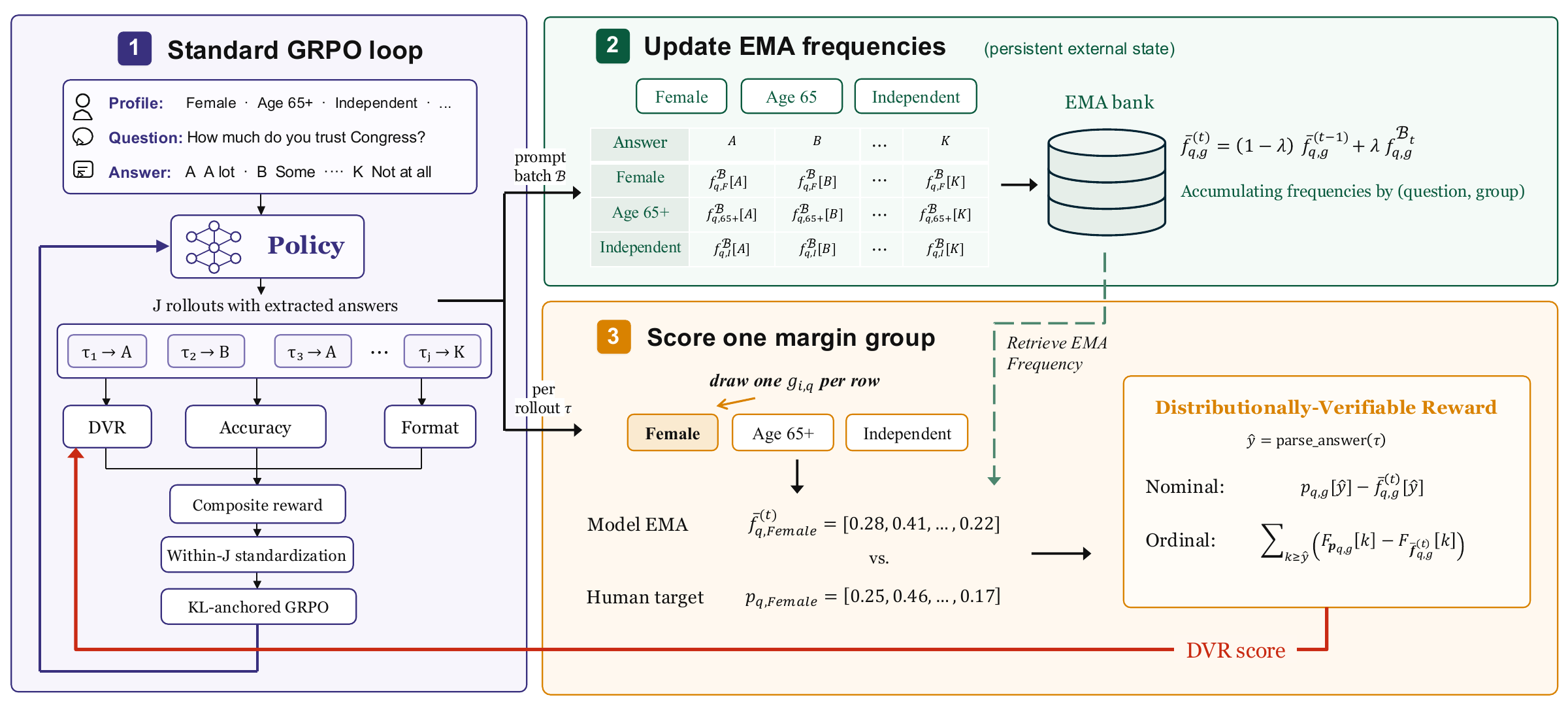}
  \caption{Overview of the DVR training framework.}
  \label{fig:method}
\end{figure}

To balance per-response predictive accuracy with group-level distributional fit, we propose \textbf{Distributionally-Verifiable Rewards (DVR)}, a GRPO-compatible reward that aligns generated responses with distributions on demographic group marginals. During training, DVR maintains online estimates of the response frequencies induced by the current policy over demographic margins. It converts discrepancies between these estimates and the corresponding human frequencies into per-completion rewards, allowing the distributional objective to be optimized through response-level RL and combined with rewards such as accuracy. Figure~\ref{fig:method} summarizes the method.

We evaluate DVR on OpinionQA \citep{santurkar2023opinionqa} with Qwen3-4B \citep{yang2025qwen3}, against accuracy-only RLVR and forward-KL distribution fitting (KL-SFT). Compared with RLVR, DVR yields higher mean accuracy in-distribution and slightly lower mean accuracy on held-out questions, \(34.5\%\) and \(14.6\%\) lower JSD in-distribution and on held-out questions respectively, and higher mean DPC point estimates on both splits. KL-SFT achieves low JSD, but its mean DPC is \(0.077\), below the zero-shot value of \(0.218\),
showing that aggregate fit alone does not establish demographic fidelity. Together, these results suggest that DVR can integrate demographic-margin objectives into response-level RL at comparable point-estimate accuracy, while underscoring the need to evaluate accuracy, distributional fit, and demographic variation jointly.

\section{Methodology}

\subsection{Problem setup}

Let $\mathcal I$ and $\mathcal Q$ denote the sets of respondents and survey questions. The observed respondent--question pairs are $\mathcal N:=\{(i,q):i\in\mathcal I,\ q\in\mathcal Q,\text{ and respondent }i \text{ was asked question }q\}$. We suppress split superscripts, so $\mathcal N$ and its derived quantities refer to the relevant split in context. Each question $q$ has a finite answer set $\mathcal K_q=\{1,\ldots,K_q\}$, and $y_{i,q}\in\mathcal K_q$ is the recorded response for $(i,q)\in\mathcal N$.

Respondent $i$ has demographic profile
$\boldsymbol a_i=(a_i^{(m)})_{m\in\mathcal M}$, where $\mathcal M$ indexes age,
gender, education, political affiliation, and other attributes.
Attribute $m$ has finite support $\mathcal A^{(m)}$; continuous attributes are
discretized into meaningful bins. A group is specified by
$\mathcal M_g\subseteq\mathcal M$ and
$g\in\prod_{m\in\mathcal M_g}\mathcal A^{(m)}$, with $i\in g$ iff $\left.\boldsymbol a_i\right|_{\mathcal M_g}=g$, e.g.,
$\mathcal M_g=\{\text{party}\}$ permits $g=\text{Republican}$. Since groups across attribute subsets overlap, a respondent may belong to several groups in the retained family $\mathcal G$.

The model induces a completion policy $\pi_{\boldsymbol\theta}$. For row
$(i,q)$, prompt $x_{i,q}$ contains the question, options, and demographics. We
draw $J$ independent rollouts
$\tau_{i,q,j}\sim\pi_{\boldsymbol\theta}(\cdot\mid x_{i,q})$ for
$j\in[J]:=\{1,\ldots,J\}$, and a fixed parser extracts
$\hat y_{i,q,j}\in\mathcal K_q$ from each parseable completion. We optimize
$\pi_{\boldsymbol\theta}$ with RL to balance per-response accuracy and
group-level distributional alignment.

\subsection{Distributionally-Verifiable Reward}
\label{sec:dvr}

\textbf{Distribution reward.} 
For a question--group pair $(q,g)$, define its respondents as
$\mathcal I_{q,g}:=\{i:(i,q)\in\mathcal N,
\left.\boldsymbol a_i\right|_{\mathcal M_g}=g\}$. The human reference
distribution is the empirical histogram
$\boldsymbol p_{q,g}:=(p_{q,g}[k])_{k\in\mathcal K_q}$, where
$p_{q,g}[k]:=|\mathcal I_{q,g}|^{-1}\sum_{i\in\mathcal I_{q,g}}
\mathbf 1\{y_{i,q}=k\}$.  We use overlapping margins because pooling erases demographic variation, whereas full demographic intersections are too sparse to estimate reliably.

For batch $\mathcal B\subseteq\mathcal N$, define the generated frequency vector
$\boldsymbol f_{q,g}^{\mathcal B}:=(f_{q,g}^{\mathcal B}[k])_{k\in\mathcal K_q}$.
Every parsed rollout contributes to every retained group matched by its respondent:
\begin{align}
f_{q,g}^{\mathcal B}[k]
=
\frac{\left|\left\{(i,j):(i,q)\in \mathcal B,\,j\in[J],\,\hat y_{i,q,j}=k,\, \left.\boldsymbol a_i\right|_{\mathcal M_g}=g\right\}\right|}{\left|\left\{(i,j):(i,q)\in \mathcal B,\,j\in[J],\,\left.\boldsymbol a_i\right|_{\mathcal M_g}=g \right\}\right|
}.
\end{align}
To reduce finite-rollout noise, we initialize an EMA on the first visit to each question--group pair and update it on subsequent visits as 
$\bar{\boldsymbol f}_{q,g}^{(t)}:=(\bar f_{q,g}^{(t)}[k])_{k\in\mathcal K_q}$. For observed $(q,g)$ in the $t$-th batch $\mathcal B_t$, we initialize $\bar{\boldsymbol f}_{q,g}^{(1)}=\boldsymbol f_{q,g}^{\mathcal B_1}$ and update
\begin{align}
  \bar{\boldsymbol{f}}_{q,g}^{(t)} &=(1-\lambda)\bar{\boldsymbol{f}}_{q,g}^{(t-1)}+\lambda \boldsymbol{f}_{q,g}^{\mathcal B_t}.
\end{align}
After updating every retained group matched in $\mathcal B_t$, for each
$(i,q)\in\mathcal B_t$ we uniformly draw one membership $g_{i,q}$ of respondent $i$. The draw selects only the residual used to score that row's $J$ rollouts, not the groups used for frequency estimation. Nominal answers receive the signed residual at their answer coordinate; ordinal answers receive the cumulative residual. For $\boldsymbol d:=(d[k])_{k\in\mathcal K_q}$, define $F_{\boldsymbol d}(k):=\sum_{\ell\in\mathcal K_q:\,\ell\le k}d[\ell]$. Then
\begin{equation}
R^{\mathrm{dist}}_{i,q,j}=
\begin{cases}
p_{q,g_{i,q}}[\hat y_{i,q,j}] - \bar f_{q,g_{i,q}}^{(t)}[\hat y_{i,q,j}], & \text{nominal}, \\
\sum_{k\ge\hat y_{i,q,j}}
\left(F_{\boldsymbol p_{q,g_{i,q}}}(k)-F_{\bar{\boldsymbol f}_{q,g_{i,q}}^{(t)}}(k) \right), & \text{ordinal}.
\end{cases}
\end{equation}

\textbf{Auxiliary rewards.}
For parsed answers,
$R^{\mathrm{acc}}_{i,q,j}=\mathbf 1\{\hat y_{i,q,j}=y_{i,q}\}$;
unparseable outputs are $R^{\mathrm{acc}}_{i,q,j}=-0.5$.
The format reward $R^{\mathrm{fmt}}_{i,q,j}$ is $1$ if the completion follows the expected \texttt{<reasoning>}--\texttt{<answer>} structure and contains a parseable answer, and $0$ otherwise. We combine the rewards as
\begin{equation}
R_{i,q,j}
= \alpha_1 R^{\mathrm{dist}}_{i,q,j}
+ \alpha_2 R^{\mathrm{acc}}_{i,q,j}
+ \alpha_3 R^{\mathrm{fmt}}_{i,q,j},
\end{equation}
and optimize the resulting reward with standard GRPO \citep{shao2024deepseekmath}. In an idealized answer-distribution space, the distribution residual follows the negative functional-gradient direction of a squared question--group moment error.

\section{Experimental setup}

\paragraph{Data and training.}
We use OpinionQA \citep{santurkar2023opinionqa}, training on waves 26--27 and evaluating on respondent-disjoint data for the same questions (in distribution, ID) and unseen questions from wave 34 (out of distribution, OOD). Evaluation covers the overall group and first-order margins over 12 demographic attributes; full data details are in Appendix~\ref{app:experimental-details}. 
Using Qwen3-4B \citep{yang2025qwen3} as the backbone, RLVR and DVR share a completion-masked SFT initialization, prompt, parser, and GRPO setup, sampling $J=8$ responses per row at temperature \(1.0\). DVR uses $\lambda=0.2$ and reward weights $(\alpha_1,\alpha_2,\alpha_3)=(1.0, 1.0, 0.1)$. Each RL model is trained for one epoch per run, using three training seeds. Full optimization details are in Appendix~\ref{app:implementation}.

\paragraph{Baselines.} We set up a series of benchmarks for comparison.
The \textbf{blind marginal} is an untrained diagnostic that assigns every respondent the same question marginal, ignoring demographics (training-derived for ID and oracle for OOD);
\textbf{zero-shot} uses the base model with the shared prompt and parser.
\textbf{KL-SFT} adapts the forward-KL answer-distribution objective of \citet{suh2025subpop}. The answer distribution is obtained by normalizing the next-token probabilities for question options, and minimizes a forward-KL loss.
\textbf{RLVR} is the controlled accuracy-oriented comparison: it shares DVR's initialization and RL pipeline but omits $R^{\mathrm{dist}}$.
Appendix~\ref{app:comparisons} gives the exact implementations.

% respondent the same question-level answer distribution estimated from the
% training split, ignoring demographics groups. It can match pooled marginals but cannot capture systematic subgroup differences.
% \textbf{Zero-shot} uses the base model with the shared prompt and parser.
% \textbf{KL-SFT} It is intuitive to directly supervise fine tune the pred probability distribution targeting the ground truth distribution, as \citep{suh2025subpop} concluded. We use it as a representative as a distribution maxing baseline. The answer distribution is obtained by normalizing the next-token probabilities for question options, and minimizes a forward-KL loss.
% \textbf{RLVR} As proved by \citep{berdoz2026reasoning}, GRPO with accuracy reward significantly improved overall accuracy rate for human opinion learning. We adopted it as a representative as a accuracy maxing baseline, shares the same RL training process as DVR, except removing distribution reward (i.e. $\alpha_1$ = 0). 

\paragraph{Evaluation.}
All methods are evaluated from sampled responses along three complementary axes: 
\textbf{Accuracy} micro-averages exact matches over decodes;
\textbf{Jensen--Shannon divergence (JSD)} macro-averages question--group distribution gaps; 
and \textbf{distinctive profile correlation (DPC)} measures the correlation between model and human subgroup deviations from each question's Overall profile \citep{wood2016correlates}. Accuracy and DPC are
higher-better, whereas JSD is lower-better. Appendix~\ref{app:metric-implementation}
defines the sampled-response aggregation and metrics exactly.

% \paragraph{Evaluation.} We evaluate the model performance from three different perspectives with the following metrics.
% \begin{enumerate}
%     \item \textbf{Accuracy.} It measures the ultimate ability of understanding both questions and human groups. For per decode, \(\mathrm{Acc}_{i,q,j}=\Pr(\hat y_{i,q,j}=y_{i,q}\mid \hat y_{i,q,j}\in\mathcal K_q)\), and then aggregated in macro average.
%     \item \textbf{Jensen-Shannon Divergence (JSD)} This metric represents distribution fit across demographic group. Denote \(\boldsymbol f_{q,g}\) as the frequency vector on test set per question and group. With \(\boldsymbol m_{q,g}=(\boldsymbol f_{q,g}+\boldsymbol p_{q,g})/2\), \(\operatorname{JSD}_{q,g}= \tfrac12[D_{\mathrm{KL}}(\boldsymbol f_{q,g}\Vert\boldsymbol m_{q,g})+ D_{\mathrm{KL}}(\boldsymbol p_{q,g}\Vert\boldsymbol m_{q,g})]\). We macro-average it over all  question--group pairs.
%     \item \textbf{Distinctive profile correlation (DPC)} Adapting the distinctive-profile analysis of \citet{wood2016correlates}, DPC measures whether the model recovers demographic group heterogeneity instead of each question's overall. \( \mathrm{DPC}=\operatorname{corr}_{q,\,g\ne\mathrm{Overall},\,k\in\mathcal K_q} (f_{q,g}[k]- f_{q,\mathrm{Overall}}[k],\, p_{q,g}[k]-p_{q,\mathrm{Overall}}[k])\). It represents directional agreement rather than magnitude of subgroup deviations.
% \end{enumerate}

\section{Results}

\begin{table}[H]
  \caption{
  OpinionQA ID/OOD results over three training seeds for KL-SFT, RLVR, and DVR (mean; Appendices~\ref{app:implementation}
  and~\ref{app:seed-results}).
  % Results on ID and OOD (unseen questions) with OpinionQA dataset.
  Acc and DPC are higher-better; JSD is lower-better.
  Bold marks the best trained-method mean;
  \(^{\dagger}\) oracle OOD marginals unavailable in training.
  % Bold marks the best estimate among trained methods; the untrained blind-marginal is for diagnosis.
  % \(^{\dagger}\) oracle question marginals from OOD set due to absence in training.
  }
  \label{tab:main}
  \centering
  \small
  \setlength{\tabcolsep}{5pt}

  \begin{tabular}{lrrr@{\hspace{8pt}}rrr}
    \toprule
    & \multicolumn{3}{c}{ID}
    & \multicolumn{3}{c}{OOD} \\
    \cmidrule(lr){2-4}\cmidrule(lr){5-7}

    Method
      & Acc $\uparrow$ & JSD $\downarrow$ & DPC $\uparrow$
      & Acc $\uparrow$ & JSD $\downarrow$ & DPC $\uparrow$ \\
    \midrule

    \emph{Blind marginal}
      & \(0.4641\)
      & \(0.0071\)
      & \(-0.001\)
      & \(0.5132^{\dagger}\)
      & \(0.0062^{\dagger}\)
      & \(0.012^{\dagger}\) \\
    \midrule

    Zero-shot
      & \(0.3985\) & \(0.1688\) & \(0.218\)
      & \(0.5145\) & \(0.1198\) & \(0.231\) \\

    KL-SFT
      % & \(0.4712\) & \(\mathbf{0.0076}\) & \(0.089\)
      % & \(0.4620\) & \(\mathbf{0.0302}\) & \(0.080\) \\
      & \(0.4716\) & \(\mathbf{0.0075}\) & \(0.077\)
      & \(0.4588\) & \(\mathbf{0.0309}\) & \(0.062\) \\

    RLVR
      % & \(0.5216\) & \(0.1011\) & \(0.294\)
      % & \(\mathbf{0.5315}\) & \(0.1008\) & \(0.310\) \\
     & \(0.5156\) & \(0.1036\) & \(0.301\)
      & \(\mathbf{0.5350}\) & \(0.1030\) & \(0.310\) \\
    \midrule

    \textbf{DVR}
      % & \(\mathbf{0.5227}\) & \(0.0678\) & \(\mathbf{0.319}\)
      % & \(0.5312\) & \(0.0881\) & \(\mathbf{0.320}\) \\
      & \(\mathbf{0.5239}\) & \(0.0679\) & \(\mathbf{0.329}\)
      & \(0.5308\) & \(0.0879\) & \(\mathbf{0.328}\) \\

    \bottomrule
  \end{tabular}
\end{table}

\paragraph{DVR improves distributional fit at RLVR-level accuracy.}
In the controlled comparison with accuracy-only RLVR, DVR retains ID accuracy (\(0.5239\) vs.\ \(0.5156\)), reduces JSD by \(34.5\%\) (\(0.0679\) vs. \(0.1036\)), and has a higher DPC point estimate (\(0.329\) vs.\ \(0.301\)).
The same direction holds on the held-out questions: accuracy remains comparable (\(0.5308\) vs.\ \(0.5350\)), while JSD is \(14.6\%\) lower (\(0.0879\) vs.\ \(0.1030\)) and mean DPC is \(0.328\) rather than \(0.310\).
This represents that adding the distributional reward improves distribution alignment and popularity distinction of LLM without an observed accuracy cost.

\paragraph{Low JSD does not imply demographic group fidelity.}
% The untrained blind marginal attains ID JSD .0071 without using demographics, with its DPC effectively zero (\(-.001\)). This implies that reproducing the shared question marginal already yields low group-averaged JSD without recovering the demographic group direction. 
The blind marginal attains ID JSD \(0.0071\) without demographics but near-zero DPC (\(-0.001\)), showing that matching shared question marginals can yield low group-averaged JSD. Meanwhile, KL-SFT exhibits the similar pattern: despite ID JSD \(0.0075\), its DPC is \(0.077\), even below the untrained zero-shot (\(0.218\)), which implies that low JSD can reflect agreement with pooled question marginals without demonstrating recovery of demographic heterogeneity. 
On held-out questions, KL-SFT again combines low JSD with low DPC, and its accuracy falls below zero-shot (\(0.4588\) vs.\ \(0.5145\)), suggesting weak generalization for these runs. Thus, DVR offers a more balanced operating point across accuracy, distributional fit, and demographic variation rather than minimizing JSD alone.

Individual-seed results and additional topic- and demographic-group breakdowns, together with an
illustrative case study, are reported in
Appendix~\ref{app:breakdowns}.

% KL-SFT nearly reaches this marginal-only reference on ID (JSD .0076), but has DPC .089, even below zero-shot (.218). The same separation appears on held-out questions. It means that KL loss in SFT maximize the overall representation on questions, while losing population heterogeneity. 

% Moreover, note that unlike RL methods (DVR and RLVR), KL-SFT's held-out accuracy falls below the zero-shot model (.4620 vs.\ .5145), suggesting weak unseen-question generalization for this direct-fitting checkpoint. Thus, DVR's advantage is not minimizing JSD alone, but providing a more balanced operating point across the three evaluation axes.

\section{Discussion and limitations}
\paragraph{Training objectives encode different notions of fidelity.}
RLVR targets individual accuracy; KL-SFT directly fits distributions but may recover little demographic variation despite low divergence. 
DVR instead couples sampled responses through question--demographic residuals while retaining accuracy, keeping distributional correction composable with other verifiable rewards.

% Exact-match RLVR primarily targets individual predictive accuracy, whereas
% KL-SFT directly optimizes distributional agreement but can achieve low
% divergence while recovering little demographic variation. DVR instead couples
% sampled responses through question--demographic residuals while retaining the
% accuracy reward. Its role is therefore not to replace direct distribution
% fitting, but to make distributional correction composable with accuracy,
% format, and other verifiable RL objectives.

\paragraph{Population fidelity cannot be summarized by one metric.}
The blind marginal shows that low JSD without demographics does not establish subgroup recovery \citep{williams2026beyond}. DPC isolates the direction of demographic departures but not their magnitude. Accuracy, JSD, and DPC capture individual prediction, distributional fit, and demographic variation; joint reporting avoids reducing heterogeneous responses to one score and supports pluralistic evaluation of alignment.
% The blind marginal achieves very low JSD without using demographics, showing
% that marginal agreement does not establish recovery of subgroup structure
% \citep{williams2026beyond}. DPC removes the shared question marginal and
% measures the direction of demographic departures, but not their magnitude.
% Accuracy, JSD, and DPC therefore capture complementary aspects of individual
% prediction, distributional fit, and demographic variation. They should be
% reported jointly and accompanied by a demographics-blind reference.

\paragraph{Limitations and future work.}
The existing results cover one dataset and one backbone. The held-out wave tests unseen questions within the same survey, not transfer to a new population or domain. And the current DVR experiment aligns only unweighted first-order margins from finite-rollout, EMA-based estimates and does not guarantee fidelity for demographic intersections, cross-question dependencies, or the survey-weighted population. Future work should add datasets and model families; sweep reward weights; extend to joint attributes; and test population shift. Demographic agreement remains descriptive, not causal, and cannot replace human data collection.

% The current result is limited to one dataset, backbone, and training seed, which only provide metrics at point estimation without uncertainty across runs. 
% The held-out wave evaluates unseen questions within the same survey setting rather than transfer to a new population or domain.
% Also, the current experiment aligns only first-order unweighted demographic margins using finite-rollout, EMA-based estimates, which does not guarante fidelity for demographic intersections, cross-question dependencies, or the survey-weighted population. 
% Future work should repeat training across seeds, datasets, and model families; sweep the reward weights; extend group selection to joint-attribute intersections; and evaluate generalization under population shift. Demographic agreement should remain descriptive rather than causal and does not justify replacing human data
% collection with synthetic respondents.
\clearpage
\begin{ack}
This work was supported by the Tsinghua University Talent Introduction Fund [grant number 53330900126].
\end{ack}

{\small
\bibliographystyle{plainnat}
\bibliography{references}
}

\clearpage
\appendix
\section*{Appendix}
\input{appendix}

% \clearpage
% \input{checklist.tex}

\end{document}

%% file: appendix.tex
\raggedbottom

\section{Data construction and empirical targets}
\label{app:experimental-details}

\paragraph{Survey splits.}
We use OpinionQA records derived from Pew American Trends Panel waves 26, 27,
and 34. Respondents in waves 26 and 27 are assigned by a deterministic hash of
their panel identifier to an 80/10/10 train/validation/test split. The same
identifier is hashed identically in both waves, so an ID test respondent cannot
enter training through the other wave. Wave 34 is reserved for held-out-wave
evaluation and contains questions absent from training. It is therefore an
unseen-question test within the same longitudinal panel, not necessarily an
unseen-respondent test.

\begin{table}[H]
  \caption{Processed respondent--question rows. The reported models use the
  complete processed, hash-thinned ID and held-out-wave test files.}
  \label{tab:appendix-data}
  \centering
  \small
  \setlength{\tabcolsep}{5pt}
  \begin{tabular}{llrrl}
    \toprule
    Split & Wave(s) & Questions & Rows & Role \\
    \midrule
    Train      & 26, 27 & 174 & 169,125 & optimization \\
    Validation & 26, 27 & 174 &  17,050 & monitoring \\
    ID test    & 26, 27 & 174 &  17,073 & respondent-disjoint evaluation \\
    OOD test   & 34     &  65 &   6,598 & unseen-question evaluation \\
    \bottomrule
  \end{tabular}
\end{table}

Each row contains one non-refusal answer by one respondent to one question.
Rows are thinned with a question-specific hash, targeting approximately 1,000
training rows and 100 validation or test rows per question. This avoids repeatedly selecting the same respondents for every question. Table~\ref{tab:appendix-data} reports the realized counts.

\paragraph{Demographic groups.}
We use the Overall group and first-order demographic groups for region, age, sex, education, citizenship, marital status, religion, religious attendance, political party, income, political ideology, and race/ethnicity. Respondents generally belong to several groups, so memberships overlap and the resulting question--group estimates are not independent.

For ID data, a question--group pair is included when its training-side pool
contains at least 50 non-refusal responses and its survey-weighted refusal rate
is at most \(0.5\). The same included set is used for ID training and evaluation.
Because wave 34 has no training-side pool, the gate is applied within that
wave. This yields 9,040 included ID pairs and 3,519 included OOD pairs. 8,954
and 3,502, respectively, are represented by at least one processed test row
and enter the distribution metrics.

\paragraph{Target distributions and scope.}
Survey weights are used only for the retention gate. All training targets and
evaluation references reported in this paper are unweighted. For the relevant
respondent pool,
\begin{equation}
  p^{\mathrm{train}}_{q,g}[k]
  =
  \frac{\sum_i
    \mathbf 1[i\in g,\,y_{i,q}=k]}
       {\sum_i
    \mathbf 1[i\in g,\,y_{i,q}\in\mathcal K_q]}.
  \label{eq:app-target}
\end{equation}
Missing responses and refusal options are excluded before normalization.
Training targets for waves 26 and 27 use training respondents only. We denote
the analogous evaluation-reference histogram by
\(\boldsymbol p^{\mathrm{eval}}_{q,g}\): ID references use all available
respondents in waves 26 and 27, whereas OOD references use all available
respondents in wave 34. Thus, these targets describe empirical response
distributions in the observed panel sample, not survey-weighted estimates for
the U.S.\ population.

\section{Prompts and implementation}
\label{app:implementation}

\subsection{Prompt and output contracts}

Zero-shot, the shared SFT initialization, RLVR, and DVR use the same compact
two-message prompt. Attribute values are copied from the survey data, and
missing attributes are omitted. The complete template is:
\begin{center}
\begingroup
\setlength{\fboxsep}{8pt}
\setlength{\fboxrule}{0.6pt}
\fbox{%
\begin{minipage}{\dimexpr\linewidth-2\fboxsep-2\fboxrule\relax}
\raggedright\small
\colorbox{black!8}{\parbox{\dimexpr\linewidth-2\fboxsep\relax}{\bfseries Survey respondent simulation prompt}}
\par\medskip
\textbf{System}\par\smallskip
You are a digital twin of the U.S. survey respondent described below. Reason,
then answer as that person.
\par\medskip
{\color{black!25}\hrule height 0.4pt}
\medskip
\textbf{User}\par\smallskip
\textit{U.S. survey respondent:}\par\smallskip
\begin{tabular}{@{}p{.48\linewidth}@{\hspace{.04\linewidth}}p{.48\linewidth}@{}}
Region of residence: \{region\}; & Age: \{age\}; \\
Sex: \{sex\}; & Education: \{education\}; \\
Citizenship: \{citizenship\}; & Marital status: \{marital status\}; \\
Religion: \{religion\}; & Religious attendance: \{attendance\}; \\
Political party: \{party\}; & Family income: \{income\}; \\
Political ideology: \{ideology\}; & Race/ethnicity: \{race\}
\end{tabular}
\par\medskip
\textbf{Question:} \{question text\}\par\smallskip
\begin{tabular}{@{}r@{\quad}l@{}}
1. & \{option 1\} \\
2. & \{option 2\} \\
& \ldots \\
K. & \{option K\}
\end{tabular}
\par\medskip
\colorbox{black!4}{\parbox{\dimexpr\linewidth-2\fboxsep\relax}{%
Reason, then answer with one of [1, 2, \ldots, K]:\\[.4ex]
\texttt{<reasoning>...</reasoning><answer>1</answer>}}}
\end{minipage}%
}
\endgroup
\end{center}
The displayed tags define explicit reasoning and answer fields. The parser
first reads the last \texttt{<answer>} span and otherwise falls back to the last
in-range option index in the completion.

KL-SFT uses a direct-answer contract because its loss supervises one
next-token position rather than a reasoning trace. Its system message is
``You are a digital twin of the U.S. survey respondent described below. Answer
as that person.'' The profile, question, and options are rendered as above,
followed by:
\begin{quote}
\raggedright\small
Answer with exactly one of [1, 2, \ldots, K] inside
\texttt{<answer></answer>} tags and nothing else---do not explain, e.g.
\texttt{<answer>1</answer>}.
\end{quote}
Training appends \texttt{<answer>} to the templated prompt and reads the logits
at the next-token position. Evaluation supplies the same prefix before
decoding, so generation begins at the answer-index position.

\subsection{Training seeds and checkpoint selection}
\label{app:training-seeds}

KL-SFT, RLVR, and DVR each use training-seed configuration values
\(0,1,2\). We keep each method's original data, training recipe, and terminal
checkpoint rule, changing only \texttt{data.train\_seed} and run/output names
for the two additional runs. RLVR and DVR reuse the same SFT initialization
and frozen reference across all runs. The SFT warm-up is not retrained per seed.
We evaluate the final checkpoint at 2,628 optimizer steps for KL-SFT and
21,140 steps for RLVR/DVR, without selecting checkpoints by validation scores.
The runs vary the training seed exposed by the existing code. KL-SFT uses this seed for training and row-keyed membership selection, with its LoRA model constructed before the Trainer sets the seed. DVR uses a separate membership-draw random-number generator fixed at seed 0. Each checkpoint receives one complete ID/OOD evaluation with the same rows,
references, and method-specific decoding contract, using \(J=8\) and
temperature \(1.0\) with the existing decoding-seed settings.
Metrics are computed separately for each checkpoint, then summarized as the
mean and sample SD (\(n=3\), denominator \(n-1\)), without pooling decodes across
seeds. Appendix~\ref{app:seed-results} lists the individual results.

\subsection{Shared SFT initialization}

RLVR and DVR start from the same completion-only SFT checkpoint, with prompt
tokens masked from the loss. A deterministic shuffle selects 60,000 training
rows, for which Qwen3-4B generates four completions each at temperature \(1.0\). We
keep at most the first completion whose parsed answer matches the recorded
response, obtaining 33,045 examples (\(55.1\%\)). LoRA SFT uses effective batch
size 8 for 2,300 optimizer steps (18,400 examples, or \(0.56\) epochs over the
included set), with learning rate \(5\times10^{-5}\), a cosine schedule, and 80
warm-up steps. The merged checkpoint
initializes both RLVR and DVR and is also their frozen KL reference. Zero-shot
uses the original Qwen3-4B checkpoint and is therefore not an
initialization-matched ablation.

\subsection{Exact DVR update}

Algorithm~\ref{alg:dvr} records the implemented order of operations. Every
parsed rollout first updates every included membership of its respondent.
Only after all touched EMAs have been updated do we draw one membership for a
row and use it to score that row's \(J\) rollouts. Thus the random draw controls
scoring, not frequency estimation.

\begin{algorithm}[t]
\caption{DVR training with random-margin residual rewards}
\label{alg:dvr}
\begin{algorithmic}[1]
\Require policy \(\pi_\theta\), frozen reference \(\pi_{\mathrm{ref}}\),
targets \(\boldsymbol p^{\mathrm{train}}_{q,g}\), memberships
\(\mathcal M(i,q)\), EMA rate
\(\lambda\), and \(J\) rollouts
\State leave \(\bar{\boldsymbol f}_{q,g}\) undefined for every included pair
\For{each generation batch \(B\)}
  \For{each row \((i,q)\in B\)}
    \State sample \(J\) completions and parse \(\hat y_{i,q,1:J}\)
  \EndFor
  \For{each parsed \(\hat y_{i,q,j}\) and each \(g\in\mathcal M(i,q)\)}
    \State add one count for \(\hat y_{i,q,j}\) to pair \((q,g)\)
  \EndFor
  \For{each pair \((q,g)\) with a parsed answer in \(B\)}
    \State normalize its counts to obtain \(\boldsymbol f^B_{q,g}\)
    \If{this is the pair's first visit}
      \State \(\bar{\boldsymbol f}_{q,g}\gets\boldsymbol f^B_{q,g}\)
    \Else
      \State \(\bar{\boldsymbol f}_{q,g}\gets
      (1-\lambda)\bar{\boldsymbol f}_{q,g}
      +\lambda\boldsymbol f^B_{q,g}\)
    \EndIf
  \EndFor
  \For{each row \((i,q)\in B\)}
    \State draw \(g_{i,q}\sim\operatorname{Unif}(\mathcal M(i,q))\)
    \For{\(j=1,\ldots,J\)}
      \If{\(\hat y_{i,q,j}\) is unparseable}
        \State \(R^{\mathrm{dist}}_{i,q,j}\gets-1\)
      \Else
        \State compute the nominal or ordinal residual with the updated
        \(\bar{\boldsymbol f}_{q,g_{i,q}}\)
      \EndIf
      \State \(R_{i,q,j}\gets R^{\mathrm{acc}}_{i,q,j}
      +0.1R^{\mathrm{fmt}}_{i,q,j}+R^{\mathrm{dist}}_{i,q,j}\)
    \EndFor
    \State standardize \(R_{i,q,1:J}\) within the rollout group
  \EndFor
  \State apply clipped, KL-regularized GRPO updates over the buffered microbatches
\EndFor
\end{algorithmic}
\end{algorithm}

Only parsed outputs enter the numerator and denominator of
\(\boldsymbol f^B_{q,g}\). A pair with no parsed output in a batch is not updated. The first valid visit initializes the EMA directly, and later visits use \(\lambda=0.2\). The index \(t\) in Section~\ref{sec:dvr} counts visits to a particular question--group pair, not global optimization steps. The same batch updates the
EMA before its rollouts are scored against that post-update value. Absent pairs keep their previous state.

The generation batch contains 16 prompt rows, while the training data contain 174 questions. Same-question rows therefore rarely co-occur. Under the same batching, a zero-weight diagnostic measured a sample-weighted mean of \(1.038\) distinct batch respondents sharing a group, averaged over each row's memberships. Cross-respondent aggregation is therefore supplied primarily by the EMA across visits rather than by the current batch alone.

\subsection{Rewards and GRPO optimization}

The accuracy reward is exact match for nominal and ordinal questions:
\begin{equation}
R^{\mathrm{acc}}_{i,q,j}=
\begin{cases}
1, & \hat y_{i,q,j}=y_{i,q},\\
0, & \hat y_{i,q,j}\in\mathcal K_q\ \text{and}\
     \hat y_{i,q,j}\ne y_{i,q},\\
-0.5, & \hat y_{i,q,j}\ \text{is unparseable}.
\end{cases}
\end{equation}
The format reward is one when the completion has exactly one opening
\texttt{<reasoning>} tag and one opening \texttt{<answer>} tag, contains both
closing tags, places the reasoning opening before the answer opening, and
contains a parseable option index. Otherwise it is zero. DVR uses
\(R^{\mathrm{acc}}+0.1R^{\mathrm{fmt}}+R^{\mathrm{dist}}\), whereas RLVR
removes the final term. The components are combined at their raw scales.

The composite reward is centered and divided by its sample standard deviation
within each \(J=8\) rollout group:
\begin{equation}
A_{i,q,j}=
\frac{R_{i,q,j}-J^{-1}\sum_{j'=1}^J R_{i,q,j'}}
     {\operatorname{sd}(R_{i,q,1:J})+10^{-4}}.
\end{equation}
A rollout group with identical composite rewards therefore has zero
reward-driven advantage even if its aggregate margin is misaligned. The KL
term still anchors it to the reference policy. Training uses DAPO
completion-token normalization, token-level policy importance ratios, clipping
at \(0.2\), and KL coefficient \(0.02\).

\begin{table}[H]
\caption{Configuration shared by the reported RLVR and DVR models.}
\label{tab:app-training-config}
\centering
\small
\begin{tabular}{@{}p{.28\linewidth}p{.66\linewidth}@{}}
\toprule
Item & Value \\
\midrule
Backbone & Qwen3-4B, bfloat16, SDPA attention, no quantization \\
LoRA & rank 32, \(\alpha=32\), dropout \(0.05\), all attention and MLP projections \\
Sampling & \(J=8\), temperature \(1.0\), at most 256 completion tokens, explicit
tagged reasoning--answer output \\
Batching & 8 completion sequences per training microbatch, 8 gradient-accumulation
steps (64 completions, or 8 prompts, per optimizer step), 128 completions
(16 prompts) per generation batch \\
Optimization & 21,140 optimizer steps, constant learning rate \(5\times10^{-6}\),
AdamW \((\beta_1,\beta_2)=(0.9, 0.99)\), weight decay \(0.1\), maximum gradient norm 1 \\
GRPO & clipping threshold \(0.2\), KL coefficient \(0.02\), colocated vLLM generation \\
DVR & EMA rate \(0.2\), uniform one-membership draw, reward weights
accuracy/format/distribution \(=1.0/0.1/1.0\) \\
Data and seeds & 169,125 training rows per run, approximately one epoch, seeds 0, 1, 2 \\
Pinned software & PyTorch 2.10.0, TRL 1.5.1, Transformers 4.57.6, PEFT 0.19.1,
vLLM 0.19.0 \\
\bottomrule
\end{tabular}
\end{table}

\section{Comparison-method details}
\label{app:comparisons}

\paragraph{KL-SFT.}
KL-SFT adapts the forward-KL option-distribution objective of
\citet{suh2025subpop}. For each training row, one included membership
\(g^\star_{i,q}\) is selected uniformly and deterministically from its row
identifier. Let \(z_{\theta,k}(x_{i,q})\) be the next-token logit for option
\(k\) immediately after the \texttt{<answer>} scaffold. The loss uses the
option-restricted normalization
\begin{equation}
  \widetilde{\pi}_{\theta}(k\mid x_{i,q})
  =
  \frac{\exp z_{\theta,k}(x_{i,q})}
       {\sum_{\ell\in\mathcal K_q}\exp z_{\theta,\ell}(x_{i,q})},
  \qquad
  \mathcal L_{\mathrm{KL\text{-}SFT}}
  =D_{\mathrm{KL}}\!\left(
    \boldsymbol p^{\mathrm{train}}_{q,g^\star_{i,q}}
    \,\Vert\,
    \widetilde{\boldsymbol\pi}_{\theta}(\cdot\mid x_{i,q})
  \right).
\end{equation}
It starts directly from Qwen3-4B and trains for one epoch using the same LoRA
geometry, learning rate \(5\times10^{-5}\), a cosine schedule, and 80 warm-up
steps. Integer answers 10 and 11 are multi-token under this readout, so 994 of
169,125 training rows (\(0.59\%\)) are excluded. Evaluation still covers all
questions.

At evaluation, we prefill \texttt{<answer>} and sample a direct-answer
continuation at temperature \(1.0\), allowing up to 12 tokens so multi-token option
indices remain decodable. The resulting answers pass through the same \(J=8\)
sampled-response metric implementation as the other trained methods. The table
therefore evaluates generated responses rather than reading token probabilities
analytically. Because KL-SFT differs in initialization, prompt, objective, and
optimization schedule, it is a method-family comparison rather than a
single-variable ablation.

\paragraph{Blind marginal.}
For an ID test row, this untrained diagnostic draws \(J=8\) pseudo-responses
independently from the question's Overall histogram estimated from training
respondents, \(\boldsymbol p^{\mathrm{train}}_{q,\mathrm{Overall}}\). Every
respondent receives the same distribution, so demographics are ignored.
Wave-34 questions have no training marginal. The OOD diagnostic instead draws
from \(\boldsymbol p^{\mathrm{eval}}_{q,\mathrm{Overall}}\), estimated from the
evaluation wave itself. Those OOD entries are oracle diagnostics, not
achievable prediction results. The blind entries in Table~\ref{tab:main} are
means over five pseudo-response seeds. Their nonzero JSD reflects both finite
sampling and genuine differences between the human demographic group and
Overall distributions. Their near-zero DPC is incidental finite-sample
correlation, because the generator has no demographic signal.

\section{Metric implementation}
\label{app:metric-implementation}

Every method in Table~\ref{tab:main} is evaluated from sampled responses. For
test row \((i,q)\), let
\(
V_{i,q}:=\{j\in[J]:\tau_{i,q,j}\text{ is parseable}\}.
\)
For each represented question--group pair, taking \(\mathcal I_{q,g}\) on the
relevant test split, we compute
\begin{equation}
  \boldsymbol f_{q,g}[k]
  =
  \frac{1}{\left|\{i\in\mathcal I_{q,g}:V_{i,q}\ne\varnothing\}\right|}
  \sum_{\substack{i\in\mathcal I_{q,g}\\V_{i,q}\ne\varnothing}}
  \frac{1}{|V_{i,q}|}
  \sum_{j\in V_{i,q}}\mathbf 1[\hat y_{i,q,j}=k].
  \label{eq:eval-group}
\end{equation}
Each respondent row with at least one parseable decode has equal weight within a group, even when some decodes are unparseable. Rows with no parseable decode and pairs with no usable row are excluded. Evaluation includes every eligible membership of each respondent, using the same input rows and membership rules across methods.

\paragraph{Per-decode accuracy.}
Accuracy is micro-averaged over all parseable decodes:
\begin{equation}
  \mathrm{Acc}
  =\frac{
    \sum_{(i,q)}\sum_{j\in V_{i,q}}
    \mathbf 1[\hat y_{i,q,j}=y_{i,q}]}
  {\sum_{(i,q)}|V_{i,q}|}.
\end{equation}
Unparseable decodes are excluded from both the numerator and denominator.

\paragraph{Jensen--Shannon divergence.}
For each represented pair, define
\(\boldsymbol m_{q,g}=(\boldsymbol f_{q,g}+
\boldsymbol p^{\mathrm{eval}}_{q,g})/2\) and compute
\begin{equation}
  \operatorname{JSD}_{q,g}
  =\tfrac12D_{\mathrm{KL}}
    (\boldsymbol f_{q,g}\Vert\boldsymbol m_{q,g})
  +\tfrac12D_{\mathrm{KL}}
    (\boldsymbol p^{\mathrm{eval}}_{q,g}\Vert\boldsymbol m_{q,g}).
\end{equation}
We use natural logarithms and report the unweighted macro-average over all
represented pairs, including Overall. Each pair has the same weight regardless
of respondent count.

\paragraph{Distinctive profile correlation.}
DPC removes the model and human Overall profiles separately and then computes
one Pearson correlation over all non-Overall group--option coordinates:
\begin{equation}
  \mathrm{DPC}
  =\operatorname{corr}_{q,\,g\ne\mathrm{Overall},\,k\in\mathcal K_q}
  \left(
    \boldsymbol f_{q,g}[k]-\boldsymbol f_{q,\mathrm{Overall}}[k],
    \boldsymbol p^{\mathrm{eval}}_{q,g}[k]-
    \boldsymbol p^{\mathrm{eval}}_{q,\mathrm{Overall}}[k]
  \right).
\end{equation}
DPC is computed as a single Pearson correlation over the concatenated coordinates. The
reported correlations contain 28,110 ID and 9,597 OOD coordinates. DPC measures
agreement in the direction of demographic departures, not their magnitude.

\input{analysis_breakdown/appendix_breakdown_section}

\section{Reproducibility, scope, and responsible use}
\label{app:scope}

\paragraph{Evaluation provenance.}
Each KL-SFT, RLVR, and DVR entry in Table~\ref{tab:main} aggregates training
seeds \(0,1,2\), with one final checkpoint and one stochastic \(J=8\) evaluation
at temperature \(1.0\) per run over all processed, hash-thinned evaluation rows.
Blind marginal averages five sampling seeds, and zero-shot uses its single evaluation. The six additional training runs use the RTX PRO 6000, and their ID/OOD evaluations use the RTX 5090. These runs use the same code, data, prompts, parser, and method-specific sampling configurations as seed 0.

For the original seed-0 results, RLVR's OOD decodes were produced in a separate
evaluation job from the other OOD rows, with the same checkpoint, prompt,
parser, and sampling configuration. Repeating zero-shot in those two jobs
changed Acc and JSD by at most \(0.0005\) and DPC by approximately \(0.006\).
The topic/group breakdowns and case study in Appendix~\ref{app:breakdowns} use these seed-0 evaluations.

% \paragraph{Limitations.}
% Results from one dataset, one backbone, and one training seed do not establish
% statistical superiority; small DPC differences are point-estimate differences.
% With \(J=8\), all sampled distribution estimates remain noisy. The ID test
% labels are respondent-disjoint from training, but the human ID reference
% histograms use the full survey waves and consequently include training
% respondents. This reduces reference variance while permitting a method to
% benefit from learning the shared training marginal, which is why the blind
% diagnostic is important when interpreting low JSD.

% The OOD split tests a later wave and unseen questions, not necessarily unseen
% panelists. The analysis covers first-order demographic margins only and does
% not test intersections or causal demographic effects. The unweighted reference
% describes the observed respondent sample rather than the survey-weighted
% national population. Finally, JSD macro-averages overlapping cells, whereas
% DPC weights group--option coordinates; neither construction makes questions or
% demographic categories independent replications.

\paragraph{Data and ethics.}
We use only benchmark-released fields and do not re-identify or externally link
respondents. Demographic conditioning can reproduce stereotypes and source-data
biases. Aggregate fit neither predicts individuals nor implies causal demographic
effects. Synthetic respondents should not replace consultation or data
collection. 
%Code, configurations, evaluation scripts, and decode artifacts required to reproduce the results will be released upon acceptance.

%% file: analysis_breakdown/appendix_breakdown_section.tex
\section{Additional breakdowns and case study}
\label{app:breakdowns}

\subsection{Results across three training seeds}
\label{app:seed-results}

Table~\ref{tab:main-with-sd} reports the full results with sample SDs.
Table~\ref{tab:main} reports only the means. Table~\ref{tab:seed-results}
reports KL-SFT, RLVR, and DVR separately for training seeds \(0,1,2\),
followed by their means and sample SDs.
Seed 0 is the initial run, and seeds 1 and 2 are
additional runs under each method's unchanged recipe. Each run is scored
separately on all 17,073 ID and 6,598 OOD rows with \(J=8\) responses per row.
The seed values and checkpoint-selection rules are specified in
Appendix~\ref{app:training-seeds}.

\begin{table}[H]
  \centering
  \small
  \setlength{\tabcolsep}{5pt}
  \caption{Full OpinionQA ID/OOD results with mean $\pm$ sample SD over
  three training seeds for KL-SFT, RLVR, and DVR. Blind marginal averages
  five sampling seeds. Zero-shot uses one evaluation. Higher Acc/DPC and
  lower JSD are better. Bold marks the best trained-method mean within each
  split. \(^{\dagger}\) oracle OOD marginals unavailable in training.}
  \label{tab:main-with-sd}
  \begin{tabular}{llrrr}
    \toprule
    Split & Method & Acc $\uparrow$ & JSD $\downarrow$ & DPC $\uparrow$ \\
    \midrule
    ID & \emph{Blind marginal} & \(0.4641\) & \(0.0071\) & \(-0.001\) \\
    & Zero-shot & \(0.3985\) & \(0.1688\) & \(0.218\) \\
    & KL-SFT & \(0.4716 \pm 0.0006\) & \(\mathbf{0.0075} \pm 0.0001\) & \(0.077 \pm 0.011\) \\
    & RLVR & \(0.5156 \pm 0.0116\) & \(0.1036 \pm 0.0030\) & \(0.301 \pm 0.007\) \\
    & \textbf{DVR} & \(\mathbf{0.5239} \pm 0.0061\) & \(0.0679 \pm 0.0011\) & \(\mathbf{0.329} \pm 0.009\) \\
    \midrule
    OOD & \emph{Blind marginal} & \(0.5132^{\dagger}\) & \(0.0062^{\dagger}\) & \(0.012^{\dagger}\) \\
    & Zero-shot & \(0.5145\) & \(0.1198\) & \(0.231\) \\
    & KL-SFT & \(0.4588 \pm 0.0028\) & \(\mathbf{0.0309} \pm 0.0008\) & \(0.062 \pm 0.018\) \\
    & RLVR & \(\mathbf{0.5350} \pm 0.0038\) & \(0.1030 \pm 0.0034\) & \(0.310 \pm 0.007\) \\
    & \textbf{DVR} & \(0.5308 \pm 0.0026\) & \(0.0879 \pm 0.0025\) & \(\mathbf{0.328} \pm 0.007\) \\
    \bottomrule
  \end{tabular}
\end{table}

\begin{table}[H]
  \centering
  \small
  \setlength{\tabcolsep}{5pt}
  \caption{Individual training-seed results, means, and sample SDs ($n=3$).
  OOD denotes the unseen Wave-34 questions. Acc and DPC are higher-better.
  JSD is lower-better. Each summary is calculated from unrounded run metrics.}
  \label{tab:seed-results}
  \begin{tabular}{llrrr@{\hspace{8pt}}rrr}
    \toprule
    & & \multicolumn{3}{c}{ID} & \multicolumn{3}{c}{OOD} \\
    \cmidrule(lr){3-5}\cmidrule(lr){6-8}
    Method & Seed & Acc $\uparrow$ & JSD $\downarrow$ & DPC $\uparrow$
      & Acc $\uparrow$ & JSD $\downarrow$ & DPC $\uparrow$ \\
    \midrule
    KL-SFT & 0 & 0.4712 & 0.0076 & 0.089 & 0.4620 & 0.0302 & 0.080 \\
    KL-SFT & 1 & 0.4723 & 0.0075 & 0.073 & 0.4566 & 0.0318 & 0.060 \\
    KL-SFT & 2 & 0.4713 & 0.0074 & 0.068 & 0.4578 & 0.0306 & 0.045 \\
    KL-SFT & Mean & 0.4716 & 0.0075 & 0.077 & 0.4588 & 0.0309 & 0.062 \\
    KL-SFT & SD & 0.0006 & 0.0001 & 0.011 & 0.0028 & 0.0008 & 0.018 \\
    \midrule
    RLVR & 0 & 0.5216 & 0.1011 & 0.294 & 0.5315 & 0.1008 & 0.310 \\
    RLVR & 1 & 0.5023 & 0.1069 & 0.308 & 0.5346 & 0.1069 & 0.302 \\
    RLVR & 2 & 0.5230 & 0.1028 & 0.302 & 0.5390 & 0.1013 & 0.317 \\
    RLVR & Mean & 0.5156 & 0.1036 & 0.301 & 0.5350 & 0.1030 & 0.310 \\
    RLVR & SD & 0.0116 & 0.0030 & 0.007 & 0.0038 & 0.0034 & 0.007 \\
    \midrule
    DVR & 0 & 0.5227 & 0.0678 & 0.319 & 0.5312 & 0.0881 & 0.320 \\
    DVR & 1 & 0.5186 & 0.0668 & 0.335 & 0.5280 & 0.0854 & 0.332 \\
    DVR & 2 & 0.5305 & 0.0690 & 0.334 & 0.5332 & 0.0903 & 0.332 \\
    DVR & Mean & 0.5239 & 0.0679 & 0.329 & 0.5308 & 0.0879 & 0.328 \\
    DVR & SD & 0.0061 & 0.0011 & 0.009 & 0.0026 & 0.0025 & 0.007 \\
    \bottomrule
  \end{tabular}
\end{table}

Compared with RLVR, DVR has lower JSD and higher DPC on both splits in each
of the three corresponding runs. The individual results are reported alongside their means and sample SDs.

\paragraph{Scope of the original seed-0 breakdowns.}

The topic/group breakdowns and case study below use the seed-0 evaluations. Acc is micro-averaged over parseable decodes within each slice. JSD is macro-averaged over represented question--group pairs, and DPC is computed as one correlation over the concatenated non-Overall group--option deviations in each slice. The slices share trained models and have overlapping topic labels and demographic memberships, so their results are descriptive point estimates rather than independent replications.

\subsection{Variation across question topics}
\begin{figure}[H]
  \centering
  \includegraphics[width=\linewidth]{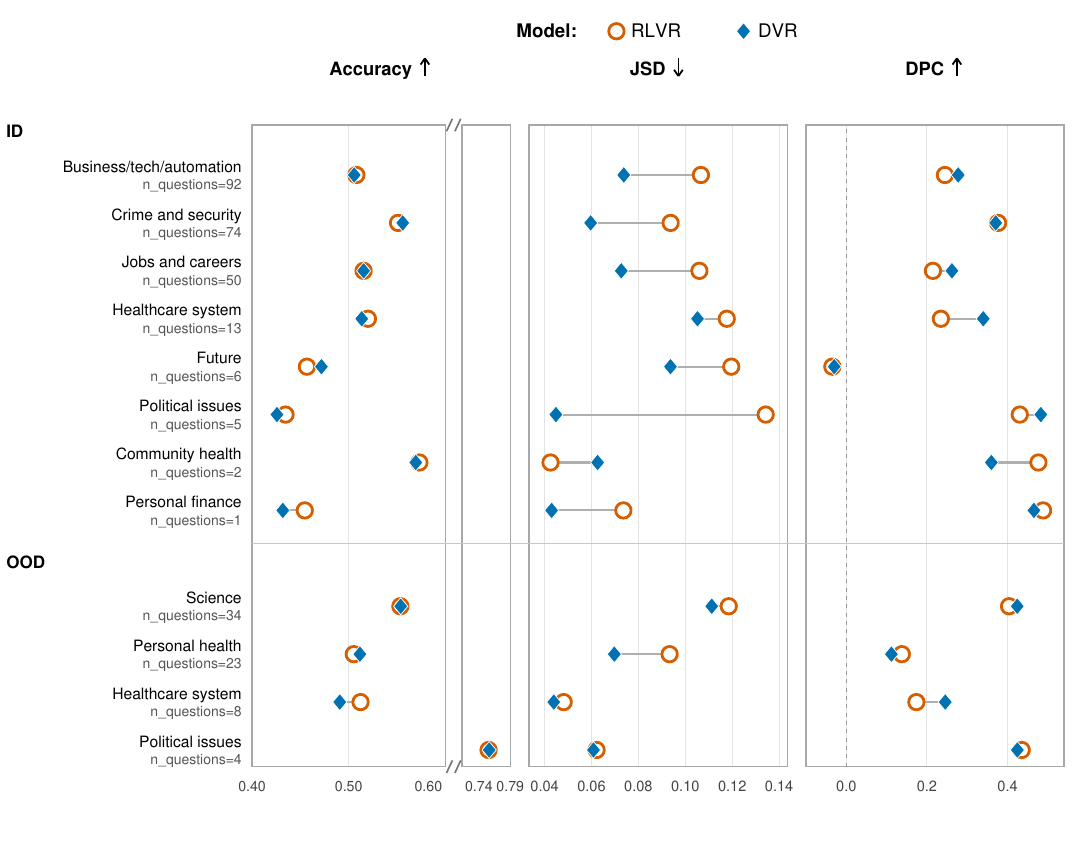}
  \caption{Topic-level RLVR--DVR comparisons on ID and held-out (OOD) questions. Rows follow OpinionQA's source-provided topic taxonomy and show the number of questions beside each topic label. Columns report absolute Acc, JSD, and DPC values, with each dumbbell connecting the two methods' estimates. Axes are independently scaled, so horizontal distances are comparable within each panel. The Accuracy axis is broken between \(0.60\) and \(0.74\), expanding the \(0.40\)--\(0.60\) range while also displaying the higher OOD political-issues values. Higher Acc and DPC and lower JSD are better.}
  \label{fig:topic-performance}
\end{figure}

We use OpinionQA's source-provided, multi-label coarse topic taxonomy
\citep{santurkar2023opinionqa}. The 174 ID questions yield 243 assignments across eight topics, and the 65 OOD
questions yield 69 assignments across four topics.
Figure~\ref{fig:topic-performance} compares RLVR and DVR within these topic
slices. The shared row labels report each topic's number of questions.

Relative to RLVR, DVR lowers JSD in \(7/8\) ID and all four OOD
topic slices. Accuracy differences change sign---DVR is higher in \(3/8\) ID
and \(3/4\) OOD slices---but the absolute difference is below one percentage
point in \(9/12\) slices, consistent with the near tie in
the seed-0 aggregate results (Table~\ref{tab:seed-results}). DPC is less uniform: DVR is higher in \(5/8\) ID and
\(2/4\) OOD slices. Aggregate and topic-level DPC are computed separately over their respective concatenated coordinates, and the multi-label topic slices overlap. These aggregation rules explain why slice-level improvement counts can differ from the aggregate comparison.

The exceptions reveal local trade-offs hidden by the aggregate operating
point. ID community health is the only JSD reversal and is worse for DVR on all
three metrics, but it contains only two questions. On the 23 OOD personal-health
questions, DVR has higher Acc (\(0.5116\) versus \(0.5053\)) and lower JSD
(\(0.0697\) versus \(0.0933\)) but lower DPC (\(0.112\) versus \(0.138\)).
Conversely, on eight OOD healthcare-system questions, DVR trades lower Acc
(\(0.4908\) versus \(0.5123\)) for lower JSD (\(0.0439\) versus \(0.0481\))
and higher DPC (\(0.246\) versus \(0.174\)). Thus, the JSD reduction is broad
across topics, whereas accuracy and demographic-direction recovery remain
topic-dependent. Slices containing only one or two questions should be treated
as question-specific observations rather than stable topic effects.

\subsection{Variation across demographic groups}
Table~\ref{tab:demographic-attributes} reports the signed change from RLVR to
DVR for 55 named groups nested within 12 demographic attributes. The numbers are absolute changes. 

\input{analysis_breakdown/demographic_attribute_comparison}

At the named-group level, JSD is the most consistent result: DVR is lower than
RLVR for all 55 ID groups and all 54 available OOD groups. Accuracy changes
straddle zero---DVR is higher in \(33/55\) ID and \(25/54\) OOD groups---and
the unweighted median signed changes are only \(+0.0010\) and \(-0.0007\),
respectively, consistent with the aggregate near tie. DPC increases in
\(36/55\) ID and \(35/54\) OOD groups, supporting the aggregate positive
direction while documenting substantial heterogeneity. At the displayed
precision, the bold rows identify Pareto improvements for \(20/55\) ID and
\(15/54\) OOD groups. The remaining rows involve at least one trade-off. JSD improves most consistently across groups, while accuracy and DPC changes vary by group.

Several systematic reversals need attention. On OOD questions, DPC decreases
for every age group (ranging from \(-0.1065\) to \(-0.0390\)) despite lower JSD
throughout. Both sex groups likewise have lower ID DPC (\(-0.0756\) for female
and \(-0.0873\) for male respondents) but higher OOD DPC (\(+0.0489\) and
\(+0.0696\)). These age- and sex-specific results show how demographic-direction recovery varies across attributes.

\subsection{Case study}

We select one question for a focused case study. It is a crime/security
question from OpinionQA's Pew American Trends Panel Wave 26
\citep{santurkar2023opinionqa}: ``How much, if at all, do you think the ease
with which people can legally obtain guns contributes to gun violence in the
country today?'' We examine its whole-question performance across all included
first-order demographic groups and its response profiles by political-party
identification.

\input{analysis_breakdown/case_analysis/joint_case_metrics}

At the whole-question scope in Table~\ref{tab:id-gun-case}, DVR improves over
RLVR by \(4.02\) percentage points in accuracy (\(0.3458\) versus \(0.3056\)),
reduces JSD by \(78.2\%\) (\(0.0293\) versus \(0.1344\)), and raises DPC from
\(0.163\) to \(0.473\). It also improves all three metrics over zero-shot.
KL-SFT achieves the lowest whole-question JSD (\(0.0174\)), but has lower Acc
(\(0.2772\)) and DPC (\(0.167\)), echoing the main-table separation between low
divergence and a more balanced operating point.

\begin{figure}[H]
  \centering
  \includegraphics[width=\linewidth]{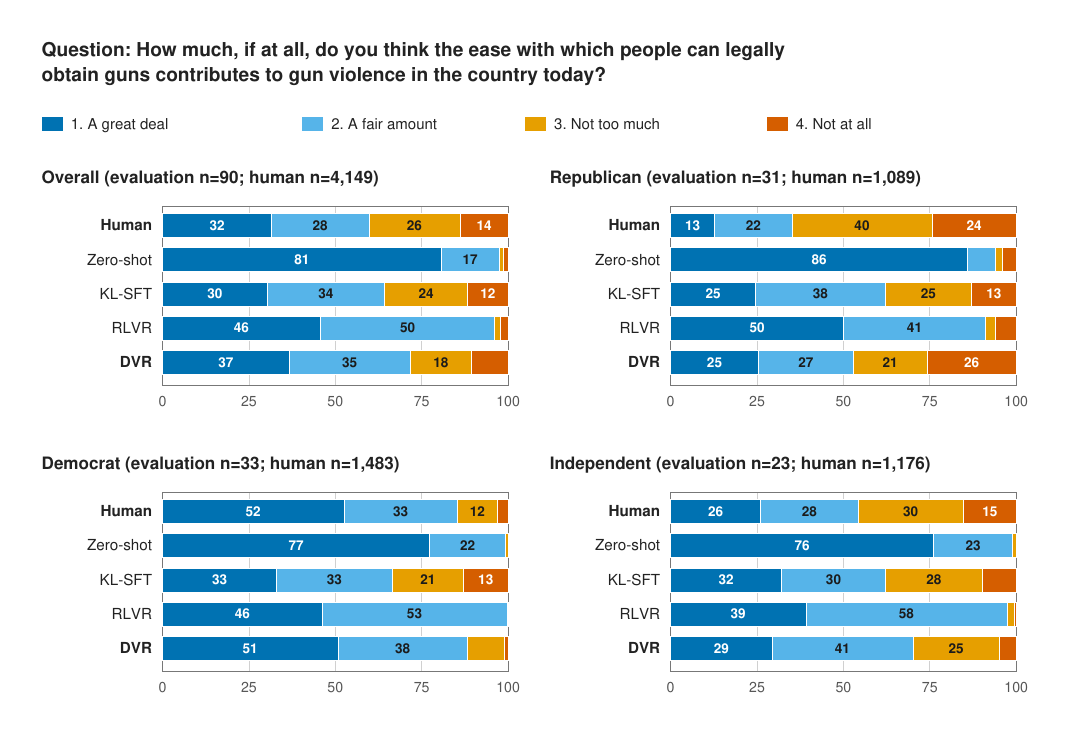}
  \caption{Response distributions for the selected OpinionQA Wave 26 ID case,
  overall and by political-party identification. Model bars pool eight samples
  per ID test respondent. Human bars use full-wave empirical references.
  Evaluation \(n\) is 31, 33, and 23 for Republican, Democrat, and Independent
  respondents, respectively. Percentages are rounded.}
  \label{fig:id-gun-profiles}
\end{figure}

Figure~\ref{fig:id-gun-profiles} shows the Overall and political-party response
profiles used to interpret this case. The whole-question metrics in
Table~\ref{tab:id-gun-case} still cover all included first-order demographic
groups.

Within the displayed political-party slice, DVR has lower JSD and higher DPC
(\(0.0173\) and \(0.865\)) than RLVR (\(0.1356\) and \(0.219\)) or KL-SFT
(\(0.0255\) and \(0.510\)). ``A great deal'' is selected by \(52\%\) of human
Democrats and \(13\%\) of human Republicans. DVR preserves this direction
(\(51\%\) versus \(25\%\)), whereas RLVR reverses it (\(46\%\) versus
\(50\%\)). The JSD ranking is scope-dependent: KL-SFT is lowest over the whole
question, but DVR is lowest within the political-party slice.

%% file: analysis_breakdown/demographic_attribute_comparison.tex
{\footnotesize
\makeatletter
\let\DVRtablelabel\fnum@table
\renewcommand{\fnum@table}{\normalsize\DVRtablelabel}
\makeatother
\setlength{\tabcolsep}{2.8pt}
\setlength{\LTcapwidth}{\linewidth}
\renewcommand{\arraystretch}{1.08}
\begin{longtable}{@{}p{.29\linewidth}rrrrrr@{}}
  \caption{Group-level changes from RLVR to DVR on ID and OOD questions. Each entry is the signed absolute change $\Delta={\rm DVR}-{\rm RLVR}$ for a named demographic group. Positive Acc and DPC changes and negative JSD changes favor DVR. Within each split, a bold triplet marks a descriptive Pareto improvement at the displayed precision: DVR is no worse on any metric and better on at least one. DPC is recomputed over the selected group's concatenated question--option deviations. The OOD Mormon group has no evaluation rows.}
  \label{tab:demographic-attributes} \\
  \toprule
  & \multicolumn{3}{c}{ID} & \multicolumn{3}{c}{OOD} \\
  \cmidrule(lr){2-4}\cmidrule(lr){5-7}
  Demographic group & $\Delta$Acc $\uparrow$ & $\Delta$JSD $\downarrow$ & $\Delta$DPC $\uparrow$ & $\Delta$Acc $\uparrow$ & $\Delta$JSD $\downarrow$ & $\Delta$DPC $\uparrow$ \\
  \midrule
  \endfirsthead
  \multicolumn{7}{c}{\tablename\ \thetable\ continued} \\
  \toprule
  & \multicolumn{3}{c}{ID} & \multicolumn{3}{c}{OOD} \\
  \cmidrule(lr){2-4}\cmidrule(lr){5-7}
  Demographic group & $\Delta$Acc $\uparrow$ & $\Delta$JSD $\downarrow$ & $\Delta$DPC $\uparrow$ & $\Delta$Acc $\uparrow$ & $\Delta$JSD $\downarrow$ & $\Delta$DPC $\uparrow$ \\
  \midrule
  \endhead
  \midrule
  \multicolumn{7}{r}{Continued on next page} \\
  \endfoot
  \bottomrule
  \endlastfoot
  \multicolumn{7}{@{}l}{\textit{Region}} \\*
  \quad Midwest & $+0.0005$ & $-0.0317$ & $-0.0440$ & $-0.0002$ & $-0.0126$ & $-0.0530$ \\
  \quad Northeast & $+0.0001$ & $-0.0316$ & $-0.1137$ & $\mathbf{+0.0018}$ & $\mathbf{-0.0121}$ & $\mathbf{+0.0163}$ \\
  \quad South & $+0.0023$ & $-0.0345$ & $-0.0364$ & $\mathbf{+0.0036}$ & $\mathbf{-0.0141}$ & $\mathbf{+0.0682}$ \\
  \quad West & $\mathbf{+0.0008}$ & $\mathbf{-0.0312}$ & $\mathbf{+0.0103}$ & $-0.0085$ & $-0.0098$ & $+0.1086$ \\
  \addlinespace[2pt]
  \multicolumn{7}{@{}l}{\textit{Age}} \\*
  \quad 18-29 & $+0.0022$ & $-0.0344$ & $-0.0784$ & $+0.0069$ & $-0.0116$ & $-0.0544$ \\
  \quad 30-49 & $-0.0003$ & $-0.0317$ & $+0.0334$ & $+0.0009$ & $-0.0158$ & $-0.0390$ \\
  \quad 50-64 & $+0.0015$ & $-0.0329$ & $-0.0547$ & $-0.0003$ & $-0.0093$ & $-0.0794$ \\
  \quad 65+ & $\mathbf{+0.0019}$ & $\mathbf{-0.0324}$ & $\mathbf{+0.0136}$ & $-0.0063$ & $-0.0123$ & $-0.1065$ \\
  \addlinespace[2pt]
  \multicolumn{7}{@{}l}{\textit{Sex}} \\*
  \quad Female & $+0.0007$ & $-0.0299$ & $-0.0756$ & $\mathbf{+0.0011}$ & $\mathbf{-0.0128}$ & $\mathbf{+0.0489}$ \\
  \quad Male & $+0.0016$ & $-0.0338$ & $-0.0873$ & $-0.0017$ & $-0.0112$ & $+0.0696$ \\
  \addlinespace[2pt]
  \multicolumn{7}{@{}l}{\textit{Education}} \\*
  \quad Associate's degree & $-0.0043$ & $-0.0355$ & $-0.0030$ & $\mathbf{+0.0142}$ & $\mathbf{-0.0143}$ & $\mathbf{+0.0530}$ \\
  \quad College graduate/some postgrad & $+0.0076$ & $-0.0345$ & $-0.0287$ & $\mathbf{+0.0014}$ & $\mathbf{-0.0159}$ & $\mathbf{+0.1384}$ \\
  \quad High school graduate & $-0.0041$ & $-0.0357$ & $+0.0853$ & $-0.0023$ & $-0.0101$ & $+0.0650$ \\
  \quad Less than high school & $\mathbf{+0.0010}$ & $\mathbf{-0.0266}$ & $\mathbf{+0.1390}$ & $\mathbf{+0.0148}$ & $\mathbf{-0.0030}$ & $\mathbf{+0.0097}$ \\
  \quad Postgraduate & $\mathbf{+0.0019}$ & $\mathbf{-0.0311}$ & $\mathbf{+0.0560}$ & $-0.0074$ & $-0.0153$ & $+0.0775$ \\
  \quad Some college, no degree & $-0.0036$ & $-0.0335$ & $+0.0318$ & $-0.0061$ & $-0.0202$ & $+0.1268$ \\
  \addlinespace[2pt]
  \multicolumn{7}{@{}l}{\textit{Citizenship}} \\*
  \quad No & $+0.0216$ & $-0.0215$ & $-0.0413$ & $\mathbf{+0.0153}$ & $\mathbf{-0.0027}$ & $\mathbf{+0.0450}$ \\
  \quad Yes & $\mathbf{+0.0012}$ & $\mathbf{-0.0323}$ & $\mathbf{+0.0289}$ & $-0.0011$ & $-0.0122$ & $+0.0639$ \\
  \addlinespace[2pt]
  \multicolumn{7}{@{}l}{\textit{Marital status}} \\*
  \quad Divorced & $+0.0065$ & $-0.0298$ & $-0.0625$ & $-0.0010$ & $-0.0154$ & $+0.0414$ \\
  \quad Married & $-0.0001$ & $-0.0341$ & $-0.0211$ & $-0.0020$ & $-0.0132$ & $+0.0468$ \\
  \quad Never been married & $-0.0020$ & $-0.0301$ & $-0.0795$ & $+0.0086$ & $-0.0108$ & $-0.0205$ \\
  \quad Separated & $\mathbf{+0.0338}$ & $\mathbf{-0.0326}$ & $\mathbf{+0.0437}$ & $-0.0154$ & $-0.0220$ & $+0.3154$ \\
  \quad Widowed & $\mathbf{+0.0092}$ & $\mathbf{-0.0271}$ & $\mathbf{+0.1041}$ & $-0.0039$ & $-0.0124$ & $-0.1969$ \\
  \addlinespace[2pt]
  \multicolumn{7}{@{}l}{\textit{Religion}} \\*
  \quad Agnostic & $\mathbf{+0.0025}$ & $\mathbf{-0.0331}$ & $\mathbf{+0.0660}$ & $-0.0027$ & $-0.0099$ & $+0.0325$ \\
  \quad Atheist & $-0.0053$ & $-0.0331$ & $+0.0531$ & $-0.0087$ & $-0.0147$ & $-0.0316$ \\
  \quad Jewish & $+0.0030$ & $-0.0352$ & $-0.0263$ & $\mathbf{+0.0016}$ & $\mathbf{-0.0163}$ & $\mathbf{+0.0646}$ \\
  \quad Mormon & $+0.0121$ & $-0.0443$ & $-0.0363$ & -- & -- & -- \\
  \quad Nothing in particular & $-0.0073$ & $-0.0310$ & $-0.0624$ & $-0.0067$ & $-0.0094$ & $+0.0183$ \\
  \quad Other & $\mathbf{+0.0111}$ & $\mathbf{-0.0237}$ & $\mathbf{+0.1085}$ & $-0.0231$ & $-0.0212$ & $-0.0399$ \\
  \quad Protestant & $+0.0065$ & $-0.0343$ & $-0.0169$ & $\mathbf{+0.0016}$ & $\mathbf{-0.0140}$ & $\mathbf{+0.0035}$ \\
  \quad Roman Catholic & $-0.0044$ & $-0.0340$ & $+0.0650$ & $+0.0027$ & $-0.0072$ & $-0.0080$ \\
  \addlinespace[2pt]
  \multicolumn{7}{@{}l}{\textit{Religious attendance}} \\*
  \quad A few times a year & $-0.0036$ & $-0.0313$ & $+0.0660$ & $+0.0001$ & $-0.0112$ & $-0.0269$ \\
  \quad More than once a week & $+0.0058$ & $-0.0379$ & $-0.0277$ & $-0.0024$ & $-0.0153$ & $-0.0238$ \\
  \quad Never & $-0.0006$ & $-0.0325$ & $+0.0049$ & $-0.0018$ & $-0.0094$ & $+0.0093$ \\
  \quad Once a week & $\mathbf{+0.0004}$ & $\mathbf{-0.0338}$ & $\mathbf{+0.0254}$ & $+0.0051$ & $-0.0143$ & $-0.0066$ \\
  \quad Once or twice a month & $\mathbf{+0.0046}$ & $\mathbf{-0.0386}$ & $\mathbf{+0.0532}$ & $-0.0153$ & $-0.0137$ & $-0.0089$ \\
  \quad Seldom & $\mathbf{+0.0024}$ & $\mathbf{-0.0306}$ & $\mathbf{+0.0156}$ & $+0.0024$ & $-0.0123$ & $-0.0210$ \\
  \addlinespace[2pt]
  \multicolumn{7}{@{}l}{\textit{Political party}} \\*
  \quad Democrat & $\mathbf{+0.0040}$ & $\mathbf{-0.0325}$ & $\mathbf{+0.0243}$ & $+0.0023$ & $-0.0152$ & $-0.0404$ \\
  \quad Independent & $-0.0019$ & $-0.0338$ & $+0.0375$ & $\mathbf{+0.0025}$ & $\mathbf{-0.0096}$ & $\mathbf{+0.0212}$ \\
  \quad Other & $-0.0021$ & $-0.0392$ & $+0.0281$ & $-0.0033$ & $-0.0106$ & $+0.0113$ \\
  \quad Republican & $\mathbf{+0.0019}$ & $\mathbf{-0.0408}$ & $\mathbf{+0.0237}$ & $-0.0063$ & $-0.0144$ & $+0.0068$ \\
  \addlinespace[2pt]
  \multicolumn{7}{@{}l}{\textit{Household income}} \\*
  \quad \$100,000 or more & $\mathbf{0.0000}$ & $\mathbf{-0.0339}$ & $\mathbf{+0.0229}$ & $\mathbf{+0.0042}$ & $\mathbf{-0.0148}$ & $\mathbf{+0.0886}$ \\
  \quad \$30,000-\$50,000 & $\mathbf{+0.0028}$ & $\mathbf{-0.0313}$ & $\mathbf{+0.0378}$ & $+0.0004$ & $-0.0152$ & $-0.1443$ \\
  \quad \$50,000-\$75,000 & $\mathbf{+0.0121}$ & $\mathbf{-0.0339}$ & $\mathbf{+0.0703}$ & $-0.0086$ & $-0.0172$ & $+0.1010$ \\
  \quad \$75,000-\$100,000 & $-0.0034$ & $-0.0341$ & $+0.0166$ & $-0.0093$ & $-0.0097$ & $+0.0832$ \\
  \quad Less than \$30,000 & $-0.0045$ & $-0.0304$ & $+0.0798$ & $\mathbf{+0.0073}$ & $\mathbf{-0.0100}$ & $\mathbf{+0.1867}$ \\
  \addlinespace[2pt]
  \multicolumn{7}{@{}l}{\textit{Political ideology}} \\*
  \quad Conservative & $-0.0009$ & $-0.0426$ & $+0.0410$ & $-0.0011$ & $-0.0101$ & $+0.0453$ \\
  \quad Liberal & $\mathbf{+0.0138}$ & $\mathbf{-0.0362}$ & $\mathbf{+0.0609}$ & $-0.0016$ & $-0.0194$ & $+0.0929$ \\
  \quad Moderate & $-0.0011$ & $-0.0329$ & $+0.0160$ & $\mathbf{+0.0019}$ & $\mathbf{-0.0115}$ & $\mathbf{+0.0492}$ \\
  \quad Very conservative & $-0.0088$ & $-0.0398$ & $+0.0236$ & $-0.0019$ & $-0.0136$ & $+0.0008$ \\
  \quad Very liberal & $-0.0024$ & $-0.0402$ & $+0.0491$ & $-0.0068$ & $-0.0166$ & $-0.0819$ \\
  \addlinespace[2pt]
  \multicolumn{7}{@{}l}{\textit{Race/ethnicity}} \\*
  \quad Asian & $\mathbf{+0.0166}$ & $\mathbf{-0.0230}$ & $\mathbf{+0.1162}$ & $+0.0034$ & $-0.0044$ & $-0.0840$ \\
  \quad Black & $-0.0052$ & $-0.0337$ & $-0.0568$ & $\mathbf{+0.0019}$ & $\mathbf{-0.0177}$ & $\mathbf{+0.0561}$ \\
  \quad Hispanic & $-0.0093$ & $-0.0262$ & $+0.0142$ & $\mathbf{+0.0075}$ & $\mathbf{-0.0087}$ & $\mathbf{+0.0758}$ \\
  \quad White & $\mathbf{+0.0014}$ & $\mathbf{-0.0326}$ & $\mathbf{+0.0071}$ & $-0.0015$ & $-0.0115$ & $+0.1012$ \\
  \addlinespace[2pt]
\end{longtable}
}

%% file: analysis_breakdown/case_analysis/joint_case_metrics.tex
% Post-hoc ID illustration. All values in this table are computed over the
% complete question, not over the political-party profiles shown in the
% accompanying figure.
\begin{table}[H]
  \caption{Whole-question performance on the selected OpinionQA Wave 26 ID case (90 test respondents). Acc pools parseable decodes. JSD is macro-averaged over the Overall and all included first-order demographic groups. DPC concatenates all non-Overall group--option departures. Bold marks the best point estimate among trained methods.}
  \label{tab:id-gun-case}
  \centering
  \small
  \setlength{\tabcolsep}{8pt}
  \begin{tabular}{lrrr}
    \toprule
    Method & Acc $\uparrow$ & JSD $\downarrow$ & DPC $\uparrow$ \\
    \midrule
    Zero-shot & \(0.2903\) & \(0.1834\) & \(0.162\) \\
    KL-SFT    & \(0.2772\) & \(\mathbf{0.0174}\) & \(0.167\) \\
    RLVR      & \(0.3056\) & \(0.1344\) & \(0.163\) \\
    \textbf{DVR} & \(\mathbf{0.3458}\) & \(0.0293\) & \(\mathbf{0.473}\) \\
    \bottomrule
  \end{tabular}
\end{table}

% The accompanying profile figure is explanatory: it is not the scope over
% which the values in this table or the question choice were computed.